# On firm specific characteristics of pharmaceutical generics and incentives to permanence under fuzzy conditions

**Javier Puente, David de la Fuente,
Jesús Lozano, Fernando Gascón***

Business Administration Department.  University of Oviedo.
jpuente@uniovi.es     david@uniovi.es
lozano@uniovi.es     fgascon@uniovi.es

### Abstract

*The aim of this paper is to develop a methodology that is useful for analysing from a microeconomic perspective the incentives to entry, permanence and exit in the market for pharmaceutical generics under fuzzy conditions. In an empirical application of our proposed methodology, the potential towards permanence of labs with different characteristics has been estimated. The case we deal with is set in an open market where global players diversify into different national markets of pharmaceutical generics. Risk issues are significantly important in deterring decision makers from expanding in the generic pharmaceutical business. However, not all players are affected in the same way and/or to the same extent. Small, non-diversified generics labs are in the worse position. We have highlighted that the expected NPV and the number of generics in the portfolio of a pharmaceutical lab are important variables, but that it is also important to consider the degree of diversification. Labs with a higher potential for diversification across markets have an advantage over smaller labs. We have described a fuzzy decision support system based on the Mamdani model in order to determine the incentives for a laboratory to remain in the market both when it is stable and when it is growing.*

**Keywords**: *Pharmaceutical generics, Permanence, entry and exit decisions; Diversification; Fuzzy decision making systems; Forecasting; Uncertainty.*

## 1    Introduction

An inevitable part of the generic drug manufacturing industry is that patents eventually become obsolete. When this occurs, the opportunity arises for third parties to exploit the market [23]. Unbranded generics start to compete with branded generics developed by a research-based pharmaceutical company. Nowadays, the industry is witnessing increased competition from generic drugs as an unprecedented number of branded medicines lose patent protection. At the same time, the drugs discovered in the labs are not replacing the value of those medicines losing patent protection [46].

Patent expirations will put at risk $267bn in drugs sales through 2016, with the largest annual spike in expirations, worth $52bn, in 2011. Pfizer's blockbuster cholesterol drug with $11bn in 2010 sales, will be the most noteworthy to go generic in 2011 [43]. As long as a gap remains between the number of drugs losing patent protection and the number of new drugs produced through R&D, it is clear that the size of the pharmaceutical industry will continue to contract. For some players, more mergers and acquisitions are likely, but others will plan to shrink, and all parts of the value chain from R&D through to production and sales and marketing will be affected [46].

Progressively, the rules of the game have been changing over the last decade. In a recent past, generic business was considered exclusively a free riding problem for pharmaceutical labs with strong R&D. However, at present, free riding activities are better regarded than in previous times and large pharmaceuticals are also considering manufacturing generics as a fundamental part of their business. Thus, the next great opportunity for traditional drugs firms is to manufacture not only generics but also biosimilars (generics which are not identical copies of biotech drugs). Although biotech-based drugs account for only a fifth or so of global drugs sales they are projected to grow at double-digit rates as sales of many conventional drugs decline. An additional factor is that the science involved in making biosimilars is much more complicated than that in making ordinary generics and this is a substantial barrier to entry for smaller labs [42].

Those generics that are manufactured by big pharmaceutical companies are often referred to as branded-generics. Branded generics are off-patent pills and potions that can be sold more cheaply than the on-patent variety, but which still command an attractive price premium mostly in poor countries due in part to the proliferation in local markets of unbranded generics which are fakes and drugs of dubious quality. In the rich world, generic drugs are advancing as a result of government action, whereas in the developing world it is the booming middle class that is propelling them forward [44].

In 2010, the big pharmaceutical companies were set to enter the branded generics markets in full force by joining up with local generics firms to get cheap access to this booming niche. Some R&D labs have already experimented with licensing deals and alliances with Asian generics firms. Others have gone a bit further by acquiring partly or even fully firms specialising in branded generics. This trend could even lead to the end of the independent generics industry in India and it is a turning-point for the global generics business [44]. Uncertainty is at its pick and so are business potential opportunities.

The real action now is in branded generics, which command a premium in many emerging markets due in part to the fear that unknown products might be fake or of dubious quality. By joining with local generics firms, multinationals can gain cheap access to the middle class in these markets. Cost-conscious governments everywhere are bashing pricey patented drugs even as they boost cheap generics [41].

In order to assess the viability of branded generics manufactured by large pharmaceuticals and non-branded generics manufactured by smaller labs, it is necessary to study the case in different regions and countries. A study in India finds that difference in price-to-patient was not as huge as it is expected for generics but margins for retailers were very high for branded-generics. The study highlights the need to modify the drug price policy and regulate the mark-ups in generic supply chain [39].

There are other recent studies that focus pharmaceutical generics. In the case of Australia, community pharmacists demonstrated a high rate of recommending generic substitution. However, to optimize the generic medicines utilization, patients' acceptance requires further improvement. Through acceptance of substitution, the patients' medicines expenditure reduced by around 21% [5].

In a study that focuses on the USA market, over 23% of physicians surveyed expressed negative perceptions about efficacy of generic drugs, almost 50% reported negative perceptions about quality of generic medications, and more than one quarter do not prefer to use generics as first-line medications for themselves or their family. Older physicians were 3.3 times more likely to report negative perceptions about generic quality, 5.8 times more likely to report that they would not use generics themselves, and 7.5 times more likely to state that they would not recommend generics for family members. Physicians reported that pharmaceutical lab representatives are the most common source of information about market entry of a generic medication [36]. These results contrast with the fact that manufacturers seeking approval to market a generic drug product must submit data demonstrating that the generic formulation is bioequivalent to the innovator drug product [7].

In a study about the Spanish market, it is found that the entry of a generic at a lower consumer price than that of the brand-name pharmaceutical or the first generic does not cause a voluntary reduction in the consumer price of either the brand drug or the first generic. Generic entry at a lower consumer price than previously existing pharmaceuticals always causes a slight reduction in the average price paid by the National Health System. The Spanish reference pricing system results in very little consumer price competition between generic firms [33].

Considering the Swedish pharmaceutical substitution reform, the price reduction due to the reform was estimated to average 10% and was found to be significantly larger for brand-name pharmaceuticals than for generics the results also imply that the reform amplified the effect that generic entry has on brand-name prices by a factor of 10. Results of demand estimation imply that the price reductions increased total pharmaceutical consumption by 8% [22].

Although economic theory indicates that it should not be necessary to intervene in the generic drug market through price regulation, most EU countries are regulating the maximum sale price of generics and the maximum reimbursement rate, especially by means of reference pricing systems. The available evidence indicates that price regulation leads to a levelling off of generic prices at a higher level than would occur in the absence of this regulation [34].

The perception of the quality of generic drugs may have to do with it [41, 44]. In Malaysia, the majority of practitioners who participated in a study claimed that they actively prescribed generic medicines in their practice. There were misconceptions among the respondents about the concepts of "bioequivalence", "efficacy", "safety", and "manufacturing standards" of generic medicines. Furthermore, advertisements and product bonuses offered by pharmaceutical companies, patient's socio-economic characteristics as well as credibility of manufacturers were factors reported to influence their choice of medicine [6].

Based only on the costs of producing generic drugs, generics could be sold for a fraction of the total patent-drug price but they end up with a far smaller discount. Thus, this potential, many times, is not realised. Once the patent expires, the patent-holder certainly loses the monopoly on the drug. However, this does not necessarily mean that there is a market environment similar to perfect competition.

Once the drug becomes off-patent, previous patent-holders change pricing policies and their marketing strategies in an attempt to retain a high market share or play even harder by thwarting competition from generics offering "pay for delay" deals that bribe rivals to put back the launch of generics. European and USA Regulators have already expressed their displeasure at perceived antitrust violations [44].

Low cost generics of patented drugs are also available in the market via agreements with the drug multinational that holds the patent, as is the case of anti-AIDS drugs in low-income countries. Some other times, patent laws are simply disregarded. The price of a year's anti-AIDS treatment was reduced dramatically by using generic drugs. Although this implies a dramatic cost reduction, this paper focuses only on non-patented generics that face generic competition.

Given the described situation, a possible alternative would be to analyse pharmaceutical generics from a macroeconomic perspective. In a complementary paper [16], we considered the macroeconomic characteristics of pharmaceutical generics in four different countries taking into account the aggregate potential not only for manufacturing, but also for consumption. Nevertheless, the current paper focuses on microeconomic business incentives.

Patents expire at different times in different countries and the percentage of medicines that are manufactured as generics vary substantially thereby making the International Trade Organisation a key player in the possible importations, exportations [15] and manufacturing of generics. Previous papers on the subject of pharmaceutical generics in one or more countries highlight both the growing interest in the study of the generics and the differences between some countries and others [8, 12, 13, 31, 38].

The European Commission's modification of the legislation on drugs and its practical implementation in each different country [35] establishes the same competition framework for generic pharmaceutical manufacturers and pharmaceutical multinationals in those cases in which the product patents have expired. The practice of country-specific price discrimination based on buying power is widely accepted in Europe for patented drugs.

In some countries, the National Health System usually negotiates a reduction in the prices of patented medicines, although eventually this might occur at the expense of the generics market. The effect that patent protection and reference prices exert on potential entrants and current stakeholders willing to enter the market of generics is also worthy of study. There is consequently a need to take into account the effect that the Health System reform and the new regulations will exert [1, 10, 14]. Our analysis focuses on the marketing of pharmaceutical generics that differently sized manufacturers put into practice, i.e., from small-scale labs to pharmaceutical multinationals. The present work describes a methodology designed to systematically study the incentives to remain in generic pharmaceutical markets.

In the 1990s, faced with new market conditions, laboratories had to refocus their competitive strategies towards increasing production efficiency or towards reinforcing R&D activities. Laboratories were used to mainly marketing two types of pharmaceutical products: in-house products (which generally provide the highest profit margins) and licensed products (developed by other labs, though still patent-protected products). Now there is a third alternative slowly gaining a market share: the marketing of generics [21].

We have briefly described the *status quo* of the competition framework in the generic pharmaceutical business: the main characteristics that affect the profit capacity and the risks involved when product patents expire. However, the complexity of changes in this field and the high number of stakeholders involved do not help to obtain a clear cut evaluation of the pros and cons of opportunities in pharmaceutical generics [29]. Pharmaceutical generic substitution rules and price regulations vary across countries and it is difficult to assess the impact of such rules on the competition framework of a specific market.

Generics are viewed as a potential solution to the problem of rising pharmaceutical costs in societies with ever-increasing health demands. In order to build a framework for the current study, after interviewing decision-makers in enterprise as well as in national and regional governments, doctors and other stakeholders, our first step was to develop a methodology to examine the decision to participate in a generic pharmaceutical market and then to apply this methodology to the Spanish market [17, 18]. Decisions to enter generic pharmaceutical markets were initially the key issue [30], although incentives to remain (entry as well as exit incentives) are now very relevant, given current uncertainties.

Although increased use of generic medicines is a way to reduce spending, uncertainty about future market size remains high. Most of the value of generics labs lies in their growth potential. Some patterns arise from different markets [45] but risks and returns are still not clear.

The aforementioned studies that compare the market situation in several countries as well as those that focus on a single country [17, 37] show the peculiarities of each country and its differences with the rest. Our methodology will make it possible to exploit this information with the aim of evaluating the economic incentives to manufacture pharmaceutical generics in a given country.

The present paper attempts to help reduce this uncertainty by intensive application of fuzzy decision support systems that take into account extensive diversification in alternative generic pharmaceutical markets.

Recent merger and takeover trends among generics labs or between R&D labs and generic firms [44, 46] are difficult to explain when the barriers to entry and exit are very low in the generic pharmaceutical business. If the barriers are very low, entry to the market is always possible without the need for an acquisition. Once in the market, there exists the need to assess whether it is worth remaining or exiting. However, this is not the case and incentives to remain have to be evaluated, which is the aim of this paper. Smaller labs interested in analysing investment opportunities in the generic pharmaceutical market do not have a clear idea of the future, but there may be incentives for large labs to build a portfolio of local generic pharmaceutical labs in different countries in order to diversify risks. Thus, diversification is the key concept in explaining recent acquisition moves in the generic pharmaceutical market.

The effect of government regulations and decisions is not easy to interpret and these represent another source of uncertainty. Risk issues are significantly important in deterring decision makers from expanding in the generic pharmaceutical business. However, not all players are affected in the same way and/or to the same extent. Small, non-diversified generics labs are in the worse position. There is, however, room for optimism: the number of generic drugs has increased in the past and will increase in the future because patent expiry is inevitable and cannot generally be postponed.

It is worth noticing again that patent expirations will put at risk $267bn in drugs sales through 2016 [43]. The exact moment a generic pharmaceutical business will take off in a given market will always be unknown, but it is nevertheless wise to be prepared. Fuzzy decision support systems can take into account uncertainty in more –and more flexible– scenarios, thereby facilitating decisions for decision makers.

Recent moves by multinational pharmaceutical labs to buy (or merge with) other generics labs cannot be explained without introducing the concept of diversification. If everyone is free to launch a generics lab, why should one company take over another one at a premium and at considerable expense, instead of simply launching a new one?

If generics can be exported and imported [25] and if competition in the market results in the penetration of pharmaceutical generics in said market in a brief period of time once the patent has expired [27], generics manufacturers should not be sold at a premium over their fixed assets value. In other words, their intangible assets should be near to zero.

Despite this apparent logic, examples of mergers abound.

Novartis is a traditionally R&D-focused Swiss pharmaceutical company who has been purchasing generic firms in the past. In 2005 it announced the purchase of Hexal, a German generics firm, and another company in the USA, Eon Labs, for a combined cost of $8.3 billion. Novartis has a generics division, called Sandoz, Novartis also paid over $8 billion in cash for these two generics labs with combined annual sales of less than $2.5 billion. Hexal and Eon would provide Novartis with more than one hundred and twenty new generic drugs and an extensive pipeline of new generic medicines [40]. These acquisition deals served to further separate Novartis from its competitors and peers. Whilst other international pharmaceutical labs still focus almost exclusively on R&D and patent-covered drugs and lobby to strengthen patent laws, Novartis has decided to bet strongly on generics, creating the world's largest manufacturer of generic drugs. Novartis also decided to acquire other labs instead of simply growing internally in the generic pharmaceutical business. Sales of biosimilars at Sandoz reached $118m in 2009 [42].

Given this scenario, it is important to reconsider the specific supply and demand characteristics of a market in order to assess not only the decisions to enter, but also those incentives to remain in or exit the generic pharmaceutical business. Moreover, given the experience and production facilities of large pharmaceutical companies, the in-house production of generics is clearly a positive net present value investment. However, this is not so clear in the case of small companies attempting to assess the incentives to remain in the market. Latent positions (firms with all the legal permits, but with very low economic activity) of small pharmaceutical labs are not strategically as viable as latent positions of larger generic pharmaceutical labs with stakes in different countries or multinational labs with relevant R&D activity that also have a significant stake in the generic pharmaceutical business.

The potential for growth in the generic pharmaceutical market is huge, despite the aforementioned uncertainties, although the potential may not be fulfilled in the near future. Pharmaceutical labs are reluctant to lose cash flows when patents expire and try to maintain their original market share.

Analysis of our results leads us to state that the option of waiting until a given generic pharmaceutical market takes off has a lower cost and higher value for well diversified pharmaceutical firms, which will condition the structure of the generic pharmaceutical industry. We expect a higher concentration and lower number of active generics labs. Large generic pharmaceutical labs are expected to enter, while small ones are expected to exit, i.e., small generic pharmaceutical labs will be picked off by larger ones.

It is interesting to evaluate generic drug opportunities in order to reduce uncertainty and to help in the decision process. This is why we are still systematically studying the viability of taking a decision to enter, be active in or exit the generic pharmaceutical market. In this paper, we model the decision to remain in or leave the market that a generics lab must take by means of a fuzzy decision support system, which is described in greater detail in the Methodology section.

The rest of the paper is structured as follows. Section 2 describes the different steps into which the methodology proposed in this paper is divided. The aim of this methodology is to quantify and limit the returns and risks that a generic pharmaceutical manufacturer has in a market. Section 3 describes how this general methodology was applied to the Spanish case analysing a number of different scenarios, while Section 4 presents the conclusions of the paper.

## 2    Methodology

The main contribution of this paper is methodological in nature. The methodology is intended to provide companies that are considering entering, staying in or exiting a market with a good decision-making framework.

We consider the Spanish case in order to show an application of the proposed methodology. Despite previous legislation in Spain (RD 2402 / 2004 & Ley 29/2006), expectations of a rapid growth of pharmaceutical generics have not materialised in practice. New legislation has been passed in 2010 (RD 4/2010 & RD 8/2010) in order to try to cut costs in the middle of a strong economic crisis and foster generics but Spain's generics labs still face high uncertainty. Pharmacies now have lower margins when dispensing generics, while reference pricing policy has not helped much. Though the methodology is applied to the Spanish market, it is also applicable to and valid for other markets.

There is no standard approach to the specific problem of determining whether entering the generic pharmaceutical business creates value or not. Our approach was to conduct several brainstorming sessions firstly to develop a Delphi questionnaire that would allow us to achieve quantitative estimates of potential sales, cost and cash flows and payback periods for an investor in generics laboratories. This data would enable us to calculate the lab's value generation. Secondly, this information would allow us to define a decision system that would make it possible to obtain the degree of incentive for staying in the generic pharmaceutical market as a function not only of the aforementioned quantitative data, but also of qualitative-type variables that influence the incentive for remaining in the market. Between each brainstorming session, the analysis of the problem was further refined to provide a consistent methodological process.

Although it is possible to evaluate the demand, i.e., the intention of consumers to buy generics [20], drugs market potential can also be evaluated from the offer side. Accordingly, a two-round Delphi questionnaire was initially designed. The corresponding questionnaires to both Delphi rounds are available on demand. The questionnaire and the kind of information that we expected to obtain are briefly described below. The questionnaire was sent to twelve Spanish manufacturers of generics. Only four of them decided to take part in the study. Although it is obvious that a higher response rate would have made it possible to obtain more accurate estimates, the methodology would have remained the same. The questionnaire included a battery of questions on the present and future status of the generic pharmaceutical market and the prospects of business opportunities in Spain. The Delphi Technique was applied in order to build up a consensus view of the subject, based upon the opinions and views of a range of experts from the field of generic medicine labs.

Respondents were told that there were no right answers to the questions posed. The goal of the research project was to provide a wide-ranging overview of the views of market-related laboratories. Average scores and typical deviations would be measured and the reasons that might lead a particular participant to hold opinions that clashed with the standard opinion of the survey would also be focused on. All participants in the survey received a second version of the questionnaire that included the average scores for each of the questions, thereby providing participants with a tool describing the opinion of a group of experts.

The aim of the first round was to obtain an initial estimation of cash flow and investment payback periods. The output of this process was a number of potential sales, costs and flow scenarios. The second round served to limit possible scenarios and to validate replies, avoiding possible contradictions between different responses. Thus, eight different scenarios in which a laboratory could compete in the generic pharmaceutical market were defined. The eight scenarios differ in terms of the number of generics, the expected sales per generic and whether there is growth in sales or not [18].

For each of the eight defined scenarios, the expected net present value (NPV) was computed through the standard expected NPV formula. However, the Net Present Value (NPV) function is difficult to evaluate because participants lack precise information. In estimating cost of capital, cash flows and payback periods, managers consider a variety of factors and their relationships. We thus adapted the first-step methodology described above to take into account the fact that participants lack precise information to estimate cash flows and payback periods while deciding how many generics to include in their portfolios and the fact that large labs are in a better position to diversify across markets than small labs.

In estimating the value of remaining in the generic pharmaceutical business, managers consider a variety of factors and the relationships between them. The NPV, the number of generics in portfolio and the degree of diversification of the lab were highlighted as the three most relevant factors. Some of these factors exhibit qualitative characteristics whose evaluation might be difficult to specify precisely (for instance, it would be more difficult to specify a lab's degree of diversification with a score of 2 on a numeric scale ranging from 1 to 10 than define it as a "low" degree of diversification among the language labels set {low, medium, high}).

Thus, with the aim of including the different factors that influence the decision of whether to stay in the generic pharmaceutical market, and given the fact that there was a certain degree of uncertainty when defining the values of some of these, fuzzy decision support systems were considered a good option to deal with the problem.

Fuzzy decision support systems are based on the theory of fuzzy sets [47] and allow an uncertainty component to be incorporated into models, making them more effective in terms of approximating to reality [26]. Linguistic variables can be used to handle qualitative or quantitative information, so that its content can be labelled taking words from common or natural language as values. This contrasts with numeric variables, which can only take numbers as values [9]. All fuzzy decision-taking problems require a knowledge base provided by an expert who is able to explain how the system works through a set of linguistic rules involving the system's input and output variables. The system's variables, i.e., the form and range of the labels for each variable, must be defined in fuzzy form. This is what Mamdani fuzzy decision support systems depend on to model systems in a process composed of five stages: the fuzzification of the input variables, the application of fuzzy operators (AND/OR) to each rule's antecedent, the implication process of each rule's antecedent to the consequent, the consequent aggregation process, and the defuzzification process [4].

The use of fuzzy decision support systems in the context of pharmaceutical generics has the advantage of allowing for flexibility in the construction of the system, swiftness in interpreting the results offered by the inference and surface maps and the ease with which a sensitivity analysis of the potential behavior of generics manufacturers under different scenarios. In order to determine the potential consumption and the potential production of pharmaceutical generics in different countries, fuzzy decision support systems were developed in a complementary paper [16] aimed at analysing the aggregate demand and the aggregate supply features of the generic pharmaceutical market from a macroeconomic perspective.

In our case, there are three input variables: the NPV, the number of generics in portfolio, and the degree of diversification. Figure 1 shows the labels and ranges defined for these variables.

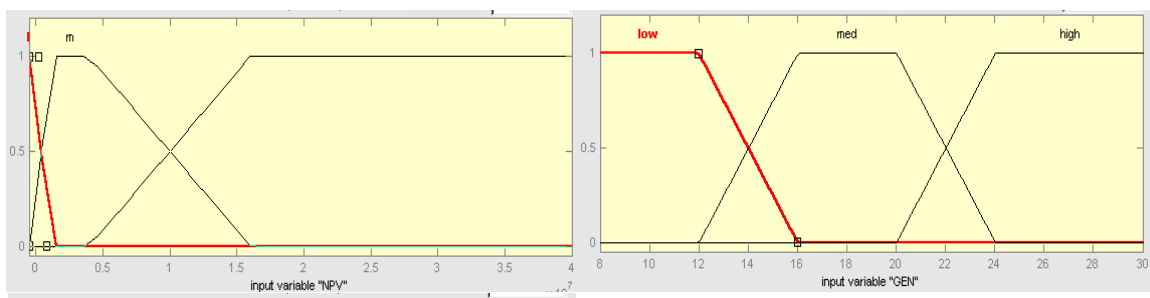

**Fig. 1:** Labels and ranges for input and output variables.

It is worth noting that the three fuzzy labels corresponding to the expected Net Present Value (NPV) variable –low, medium, high– were derived from values estimated in the two-round Delphi questionnaire, within the range (-500x10³ €, 185x10⁶ €).

The number of generics (which is an endogenous strategic variable that generics labs are able to decide upon) was separated from the expected NPV (which is an exogenous variable that depends on total market growth, government regulations and third parties decisions). In other words, generics labs were allowed to vary the number of generics in their portfolio significantly, but were unable to significantly affect expected NPV. The maximum value for the variable "number of generics" (GEN) was set to "30". Its domain was divided into three labels –low, medium, high– in a symmetric way.

We also considered the degree of diversification as a key variable. Diversification into alternative markets or being a branch of a large pharmaceutical multinational is an advantage that can only be enjoyed by large stakeholders and is unavailable to small ones. Three labels were also considered for the variable "degree of diversification" (DIVERS) –low, medium and high– which partition the range of this variable in a symmetric way. The range for this variable is measured on a scale of between 0 and 5 points.

Our system's output variable is the degree of incentive for a lab to remain (PERM-INCENT) in the generic pharmaceutical market. According to Picture 1, this variable was defined within a range measured on a scale from 0 to 100 percentage points, eight labels being defined within that range –mf1, mf2, ..., mf8– which identify different categories of growing incentive to remain in the generic pharmaceutical market.

Two fuzzy decision support systems – both based on the Mandami Model [28]– were applied in this paper. They were generated by MATLAB 6.0 – Fuzzy Logic Toolbox (v. 2.0). These systems are able to determine the incentives for a lab to remain in the generic pharmaceutical market (PERM-INCENT) on the basis of three input variables: the expected NPV of the lab (NPV), its portfolio of generics (GEN) and its degree of diversification (DIVERS). Each of these two decision support systems can be applied to its specific scenario: one to a stable market scenario and the other to a growth market scenario. Although variables were defined in the same way in the two scenarios, the output of the 27 rules proposed for collecting the knowledge of the decision differs substantially in both scenarios (self-evidently, in the market growth scenario, incentives to remain are significantly higher).

The structure of the decision system's rules is of the type: "if ((NPV is low) & (GEN is med) & (DIVERS is high)), then (PERM-INCENT is mf5)". As each one of the three input variables can take one of three categories or classes (low, medium or high), the decision framework makes use of up to 27 rules to assign the level of incentive to remain in the market (in each rule, one of eight labels can be assigned to the output variable).

Once the knowledge base is defined (entry and exit variables and decision rules), the fuzzy inference process (Mandami type) is obtained for each input tern of "crisp" values. As an example, Picture 2 depicts the inference process of the decision supplied by the system in the stable market scenario for a laboratory whose input variables values were: NPV=20x10$^6$ €, GEN=18, DIVERS=4. It can be observed that for these values the system supplies a degree of permanence in the market of 71.4% for the lab, which indicates a high degree of incentive to continue to manufacture generics in the market.

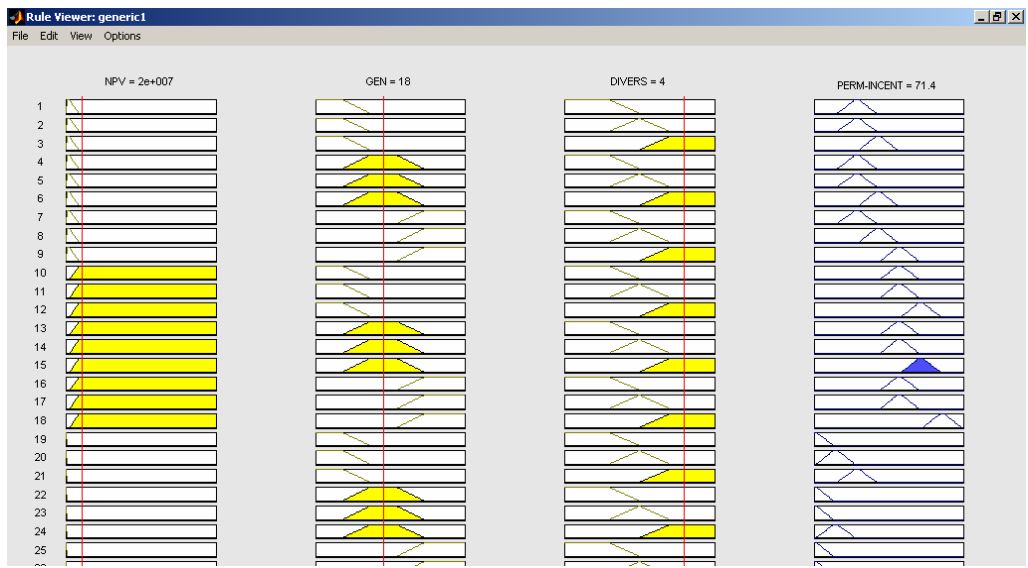

**Fig. 2:** The inference process.

A summary of inference results is presented in the next section in graphic format using the "surface viewer" of the Matlab Fuzzy Toolbox. The advantage of this way of representing the results is that the evolution that the incentive to remain undergoes can be observed as a function of any given value of two input variables for a desired specific value of the third input variable.

For instance, Picture 3 depicts the aforementioned evolution for any given value of the "GEN" and "DIVERS" variables and the specific value NPV=20x10$^6$. The interpretation of the surface is intuitive: for a high expected NPV value (20x10$^6$), incentives to stay are high regardless of the values for the other two variables; the surface starts from an incentive level of around 60% in the vertical axis. Furthermore, increasing values of "GEN" and "DIVERS", or the joint growth of both variables, result in an increase in the incentive to continue manufacturing pharmaceutical generics of up to 85%, thereby recommending any lab to remain in this context of the generic pharmaceutical market.

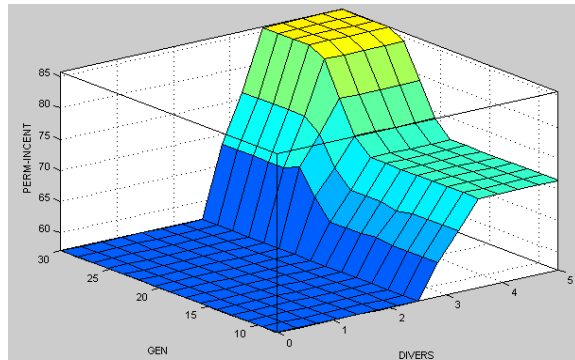

**Fig. 3:** The surface viewer.

# 3    An Application of the Fuzzy Decision Support System to the Spanish Case: Results and Discussion.

All two-by-two variable combinations for three specific values (one low, one medium, one high) of the third variable not involved in the graph were analysed in both the stable and growth scenarios. This represents the analysis of 18 surfaces. In this section, four out of these 18 surfaces are analysed since they constitute the most representative ones within the overall framework of the analysis.

Figure 4 shows the surface evolution in a stable market scenario of the incentives for a lab with an intermediate degree of diversification in alternative drug markets to remain in the market as a function of the expected NPV and the number of generics.

This surface highlights how the incentives to remain in the market analysed increase dramatically when expected NPV is very low, whatever the number of generics. Incentives to remain in the market increase to values as high as 70%. It can be seen that remaining in the market is attractive from NPV values of around 2x107 € upward, since 50% of the degree of incentive is surpassed regardless of the number of generics in portfolio. Furthermore, a sensitivity analysis of this graph shows that considering higher values in the degree of diversification and/or a market growth scenario, while keeping the same graph structure, results in a rise in the plateau value of the graph. The values for the degree of incentive are now over 85%, thereby recommending entering (or staying) in the market even more.

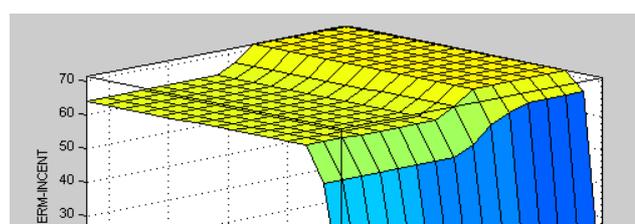

**Fig. 4:** PERM-INCENT vs. (NPV & GEN) when diversification is average (Stable Scenario).

In the stable market scenario, Figure 5 also illustrates the surface evolution of the incentives for a lab with a low expected NPV to remain in the market, in this case as a function of the level of diversification and the number of generics. In fact, the incentives to remain always increase as the degree of diversification increases, although the increase is higher for portfolios with a low number of generics. A portfolio with a high number of generics has high maintenance costs when expected NPV is low and there is no growth. Whatever the case may be, the maximum levels of incentive achieved hardly reach 30%, which is why incentives to exit are high for labs in low expected NPV cases. This situation is not sustainable in the long run unless some labs exit and some remain and acquire their rivals' market share via price wars.

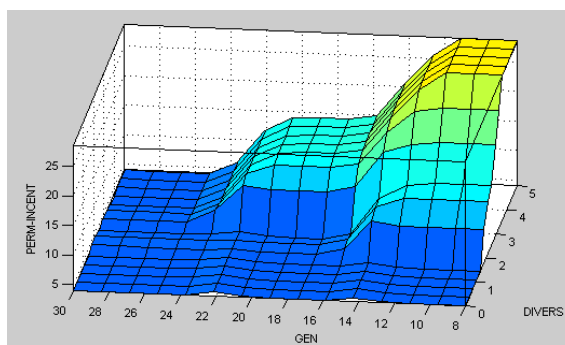

**Fig. 5:** PERM-INCENT vs. (GEN & DIVERS) when NPV is low (Stable Scenario).

Figure 6 illustrates the surface evolution of the incentives for a lab with a medium expected NPV to remain in the growth market scenario as a function of the level of diversification and the number of generics.

As the degree of diversification rises, the incentives to remain increase (up to 70%). Incentives are higher for an intermediate number of generics in the portfolio. This is because maintaining too high a number of generics is costly for the lab when diversification is low. However, if the number of generics is too low, then clients have to buy their generics from a wide range of generics labs and this is not operationally good for clients either. Although price is the key variable, clients prefer to deal with suppliers with an acceptable number of generics in their portfolio. Whatever the case may be, it is clear that potential manufacturers of generics need to count on at least a "medium" degree of diversification to obtain incentives of over 50 %, which supports the decision to enter (or remain) in the market. This policy of keeping a balanced number of generics in portfolio –the vertical intermediate plane on the graph– would achieve the maximisation of the incentive to stay as well as the recommendation to enter or remain.

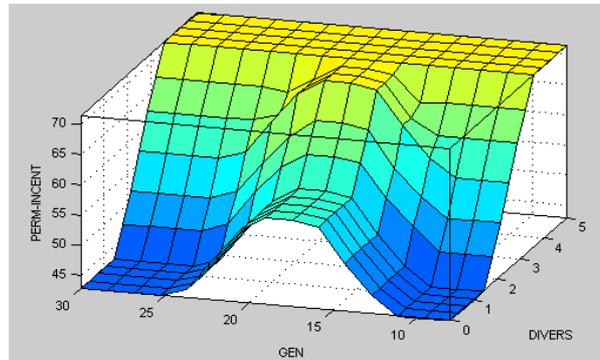

**Fig. 6:** PERM-INCENT vs. (GEN & DIVERS) when NPV is medium (Growth Scenario).

Finally, Figure 7 illustrates the surface evolution of the incentives for a firm with a high expected NPV to remain in a growth scenario as a function of the level of diversification and the number of generics.

In this scenario, all the labs should choose to increase the number of generics in their portfolio and the incentives to remain would allow all the labs to survive: the minimum incentives to remain in the market are 70% and run higher than 95%. When the level of diversification rises, the incentives to remain are higher for all cases. Thus, the recommendation under these circumstances would always be to enter or stay in the market under study, especially for laboratories with a high number of generics in portfolio and/or high degrees of diversification, since these policies would ensure high incentives to remain in the market.

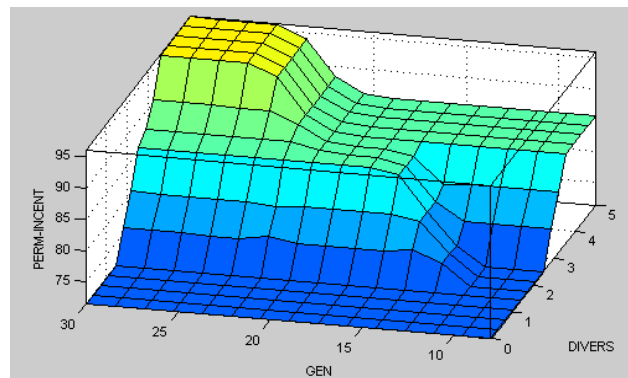

**Fig. 7:** PERM-INCENT vs. (GEN & DIVERS) when NPV is high (Growth Scenario).

The analysis of the rest of the combinations would be formulated in line with this same procedure, which is intuitive, flexible and easy to operate by any market analyst. We thus believe that the proposed decision system is sufficiently sound, since it takes into consideration the three most relevant variables in determining the incentive to remain in the generic pharmaceutical market. Furthermore, the proposed system allows the uncertainty associated with the evaluation of the variables involved in the decision model to be processed by means of the fuzzy consideration of those variables. Finally, the system is flexible in defining the labels associated with the model variables and in determining the basis for the rules that incorporate the knowledge of the decision-making process, which might vary depending on the country under study.

## 4.    Conclusions

Uncertainty is very high in the generic pharmaceutical business and each national market possesses its own peculiarities. Papers that compare different markets as well as papers that focus on a single country show the peculiarities of those markets and their differences with respect to the rest.

Given the current economic crisis, fostering generics is a clear way to cut costs and some pharmaceutical labs will profit from it but not all. High uncertainty will remain among labs. In the present paper, we have developed a methodology that enables us to exploit available firm-specific information with the aim of evaluating the economic incentives to manufacture pharmaceutical generics in a given country. We have attempted to reduce this uncertainty by systematically applying a decision tool depending on alternative scenarios. We have considered that the expected NPV and the number of generics in the portfolio of a pharmaceutical lab are important variables, but that it is also important to consider the degree of diversification. Labs with a higher potential for diversification across markets have an advantage over smaller labs. On the basis of expected NPV and the number of generics and after including diversification in the analysis, we can now better understand the economic sense of recent mergers and acquisitions in the generic pharmaceutical business.

We have described a fuzzy decision support system for an open market in which global players diversify into alternative national generics markets in order to determine the incentives for a laboratory to remain in the market both when it is stable and when it is growing. Incentives to stay in the market are significantly higher in the market growth scenario.

Once the knowledge base has been defined, the fuzzy inference process is obtained for each input tern of values (expected NPV, number of generics and degree of diversification). A summary of inference results is presented in graphic format, where incentives to remain in the market are assessed. Although this paper has focussed on the Spanish case, the methodology is valid for any pharmaceutical lab in any pharmaceutical market. Results show the consistency and adequacy of the constructed decision system to account for real decisions to enter, remain in or exit the generic pharmaceutical market. The system offers a coherent diagnosis tool of a lab's situation and also estimates how incentives fluctuate as a function of expected NPV, the number of generics and the degree of diversification.

Analysis of our results lead us to state that the option of waiting until a given generic pharmaceutical market takes off has lower cost and higher value for well diversified pharmaceutical firms, which will condition the structure of the generic pharmaceutical industry. We expect a higher concentration and lower number of active generics labs. Large generic pharmaceutical labs are expected to enter while small ones are expected to exit, i.e., small generic pharmaceutical labs will be picked off by larger ones.

The proposed methodology enables users to study the incentives a company can count on to carry out its activity in a specific generic pharmaceutical market as a function of the company's own characteristics as well as those of the market. The present analysis, which focuses on measuring the business attractiveness that a generic pharmaceutical manufacturer may have, is complemented by another study we have developed. This other study focuses on the business attractiveness that the generics market has in a country as a function of its regulation (and the way it has to put it into practice), the prices of medicines and wage costs. Although generics are a share of the total demand for pharmaceuticals, the evolution of the price of generics may be different from the evolution of the price of branded pharmaceutical products. In many countries, large pharmaceutical companies negotiate the price of branded products with governments. Sometimes, the negotiation also implies restrictions that affect the price and/or market share of generics. With these restrictions, the number of generics labs with incentives to stay in the market will be lower. In such a case, there will not be a real drop in the prices of the generic drugs; i.e., prices of generics will not be significantly lower than branded pharmaceuticals because of a lack of competition.

Finally, one issue that has not been dealt with in the present paper is the viability of state-owned generic pharmaceutical laboratories. The existence of state-owned laboratories that manufacture generic medicines usually appears in countries where the market is not the best way to allocate resources. For example, in the Soviet Union era, Bulgaria used to manufacture pharmaceutical generics for many countries within the Soviet bloc. In recent times, the Bulgarian generics business has gradually passed into private hands. In the case of some countries in Latin America or Africa, private generic pharmaceutical labs may be substituted by a state-owned laboratory which may be a feasible alternative if the bargaining power of the private laboratories is too strong.